%% file: arXiv_partialNG.tex
\documentclass[11pt]{article}
\usepackage[truedimen,margin=25truemm]{geometry}
\usepackage{microtype}
\usepackage{graphics}

\usepackage{times}
\usepackage{xcolor}
\usepackage{amsmath}
\usepackage{amsfonts}
\usepackage{amssymb}
\usepackage{url}
\usepackage{multirow}
\usepackage{mathtools}
\usepackage{booktabs}
\usepackage{graphicx}
\usepackage{epstopdf}
\usepackage{natbib}
\usepackage{algorithm2e}
\usepackage{enumerate}
\usepackage{epstopdf}

\title{E-commerce Anomaly Detection: A Bayesian Semi-Supervised Tensor Decomposition Approach using Natural Gradients}
\author{
Anil R. Yelundur$^1$, 
Srinivasan H. Sengamedu$^1$,
Bamdev Mishra$^2$\thanks{This work was carried out when the author was with India ML - Amazon.com.}%$^2$
\\ 
$^1$ India ML - Amazon.com\\
$^2$ Microsoft\\
\{yelundur,sengamed\}@amazon.com,
bamdevm@microsoft.com
}
\date{ }
\begin{document}
\maketitle
\begin{abstract}
Anomaly Detection has several important applications. In this paper, our focus is on detecting anomalies in seller-reviewer data using tensor decomposition. While tensor-decomposition is mostly unsupervised, we formulate Bayesian semi-supervised tensor decomposition to take advantage of sparse labeled data. In addition, we use P\'olya-Gamma data augmentation for the semi-supervised Bayesian tensor decomposition. Finally, we show that the P\'olya-Gamma formulation simplifies calculation of the Fisher information matrix for partial natural gradient learning. Our experimental results show that our semi-supervised approach outperforms state of the art unsupervised baselines. And that the partial natural gradient learning outperforms stochastic gradient learning and Online-EM with sufficient statistics.
\end{abstract}

\input{files/arXiv-body} % Main

\appendix

\input{files/supp_arXiv-body} % Supplementary

\bibliographystyle{named}
\bibliography{tensorcompletion}

\end{document}

%% file: files/arXiv-body.tex
\section{Introduction}

Anomaly detection implies finding patterns in the data that do not conform to normal behavior \citep{chandola09a}. In this paper, our focus is on detecting anomalies in seller-reviewer data using tensor decomposition. Anomalies in seller-reviewer graph occur when there is a secretive agreement between the sellers and reviewers for gaining an unfair market advantage. One type of abuse in Amazon e-commerce is called reviews abuse, where a seller enlists (directly or indirectly) reviewers to write fraudulent reviews about their products. In other words, sellers incentivize reviewers to create fake reviews for them and hence the major driver of reviews abuse are the sellers. In fact, there are businesses which promise reviews on Amazon for a fee. Hence detection of anomalies in seller-reviewer graph and enforcing actions on such sellers and reviewers will address the root cause of fake review origination.

From a technical perspective, we represent anomalies as \textit{dense cores} in the seller-reviewer bi-partite graph (or \textit{dense blocks} in seller-reviewer matrix) satisfying certain properties. To incorporate meta-data such as review timestamp, rating, etc. we model the data as a tensor and apply Bayesian tensor decomposition to detect anomalies. We develop semi-supervised extensions to the probabilistic tensor decomposition model to incorporate prior information regarding known bad sellers and/or reviewers. Model inference is achieved via natural gradient learning framework in a stochastic setting \citep{amari98a}.

Our contributions are as follows.
\begin{enumerate}
\item We formulate anomaly detection as a Bayesian binary tensor decomposition problem.
\item We develop semi-supervised extensions to the probabilistic binary tensor decomposition that incorporates binary target information for a subset of sellers (and/or reviewers) which have been tagged as being abusive or not abusive. This is based on the Logistic Model with P\'{o}lya-Gamma data augmentation.
\item Finally we develop partial natural gradient learning for inference of all the latent variables of the probabilistic semi-supervised model.
\end{enumerate}

Natural gradient based optimization is motivated from the perspective of information geometry and works well for many applications as an alternate to stochastic gradient descent, see {\citep{martens14a}, i.e., natural gradient learning seem to require far fewer total iterations than stochastic gradient descent. Natural gradient learning is generally applicable to the optimizaton of probabilistic models that uses \textit{natural gradient} as opposed to the standard gradient. Note that natural gradient learning requires computing the inverse of the \textit{Fisher} information matrix in every iteration. In most models, computing the Fisher information matrix is non-trivial and even if it can be computed, its inverse is usually not tractable because the size of the matrix depends on the number of latent parameters in the model. We show that the P\'{o}lya-Gamma data augmentation facilitates easy computation of the Fisher information matrix. To make the inverse of this matrix tractable, we exploit the quadratic structure of the loss function in each of the latent variables to be able to work with only the partial Fisher information matrix instead that is much smaller in size as compared with the full Fisher information matrix.

Our experiments show the following:
\begin{enumerate}
\item Our semi-supervised approach beat the state of the art unsupervised baselines in identifying abusive sellers and reviewers on Amazon data sets.
\item Partial natural gradient learning shows better learning than online EM (using sufficient statistics) on the test data set.
\item Partial natural gradient learning shows better learning than stochastic gradient learning, specifically on detecting abusive sellers in the test data set.
\end{enumerate}

The rest of this paper is organized as follows. Section \ref{sec:related_work} introduces recent related work as well as background regarding our application of tensor decomposition for anomaly detection, including our proposed partial natural gradient learning for inference. Section \ref{sec:semi_supervised} describes our proposed semi-supervised extensions to the binary tensor decomposition model via P\'{o}lya-Gamma data augmentation. Section \ref{sec:Estep} describes the modeling for all the latent variables. Section \ref{sec:Mstep} describes the inference of all the latent variables using partial natural gradients. Experimental results are shown in Section \ref{sec:experiments}. Finally, we conclude the paper in Section \ref{sec:conclusions}.

The detailed derivations of the partial Fisher information matrix as well as the gradient calculations for the proposed model are in Appendix \ref{supp_sec:fisher_matrix}.

% supplied as supplementary material available at:\\ {\color{magenta}{\url{https://dl.dropboxusercontent.com/s/jmgsow0h6zuo0hh//sub479.pdf}}}.

\section{Related Work and Background}\label{sec:related_work} 

\subsection{Fake Reviewer Detection}
There has been a lot of attention recently to address the issue of finding fake reviewers in online e-commerce platforms. Jindal and Liu \citep{jindal08a} were one of the first to show that review spam exists and proposed simple text based features to classify fake reviewers.

Abusive reviewers have grown in sophistication ever since, employing professional writing skills to avoid detection via text based techniques. To detect more complex fake review patterns, researchers have proposed graph based approaches such as approximate bi-partite cores/lockstep behavior among reviewers \citep{li16a, beutal13a, hooi16a, jiang15a}, network footprints of reviewers in the reviewer product graph \citep{ye15a} and using anomalies in ratings distribution \citep{hooi16b}.

Some recent research has also pointed out the importance of time in identifying fake reviews because it is critical to produce as many reviews as possible in a short period of time to be economically viable. Methods based on time-series analysis \citep{li15a,ye16a} have been proposed.

Tensor based methods such as CrossSpot \citep{jiang15b}, M-Zoom \citep{shin16a}, MultiAspectForensics \citep{maruhashi11a} have been proposed to identify dense blocks in tensors or dense subgraphs in heterogenous networks, which can also be applied to our case of identifying fake reviewers. Tensors are higher-dimensional generalization of matrices i.e., higher-dimensional arrays, that can seamlessly incorporate many types of information. The different dimensions of the tensor are called modes.
M-Zoom (Multidimensional Zoom) is an improved version of CrossSpot that computes dense blocks in tensors which indicate anomalous or fraudulent behavior. The number of dense blocks (i.e., subtensors) returned by M-Zoom can be configured a-priori. Note that the blocks returned may be overlapping i.e., a tuple could be included in two or more blocks. The authors in \citep{shin16a} have implemented M-Zoom and we have leveraged it as one of the baselines (un-supervised learning) in our experiments. MultiAspectForensics on the other hand automatically detects and visualizes novel patterns that include bi-partite cores in heterogenous networks.

\subsection{Anomaly Detection using Tensor Factorization}
Tensor factorization \textit{aka} decomposition can be applied to detect anomalies in seller-reviewer data because it facilitates detection of dense bi-partite subgraphs. The reasoning is as follows: at any given time, there is a common pool of fake reviewers (paid reviewers) that are available and who are willing to write a positive or negative review in exchange for a fee. Sellers recruit these fake reviewers through an intermediary channel (such as facebook groups, third party brokers, etc.) as shown in Figure~\ref{biPart:fig}; where nodes on the left indicate sellers and nodes on the right indicate reviewers and an edge indicates a written review. Note that a seller has to recruit a \textit{sizeable} number of fake reviewers in a short amount of time to make an impact on the overall product rating: positive impact for his own and negative impact for his competitor's products. Given a common pool of available fake reviewers at any given time, anomalies manifest as approximate bi-partite connections between a group of sellers and a group of reviewers with similar ratings, such as near $5$-star or near $1$-star ratings. Hence the presence of a bi-partite core (a dense bi-partite sub-graph) in some contiguous time interval $\Delta t$ is a strong indicator of anomalous connections between the group of sellers and reviewers involved.

Therefore our goal is to find bi-partite cores (or dense blocks) using tensor decomposition. Decomposing a tensor implies computing the factor matrices for each mode of the tensor. The modes of the tensor in our problem space correspond to the seller, reviewer, product, rating, and time of review. The entities in the corresponding factor matrices that have relatively higher values indicate anomalies. By aggregating these anomalous entities across the modes of the tensor results in discovering bi-partite cores, where each core consists of a group of reviewers that provide similar rating across a similar set of sellers (and their products) where all of these reviews are occuring with-in a short contiguous interval of time.

\begin{figure}[t]
\center
\includegraphics[scale=0.6]{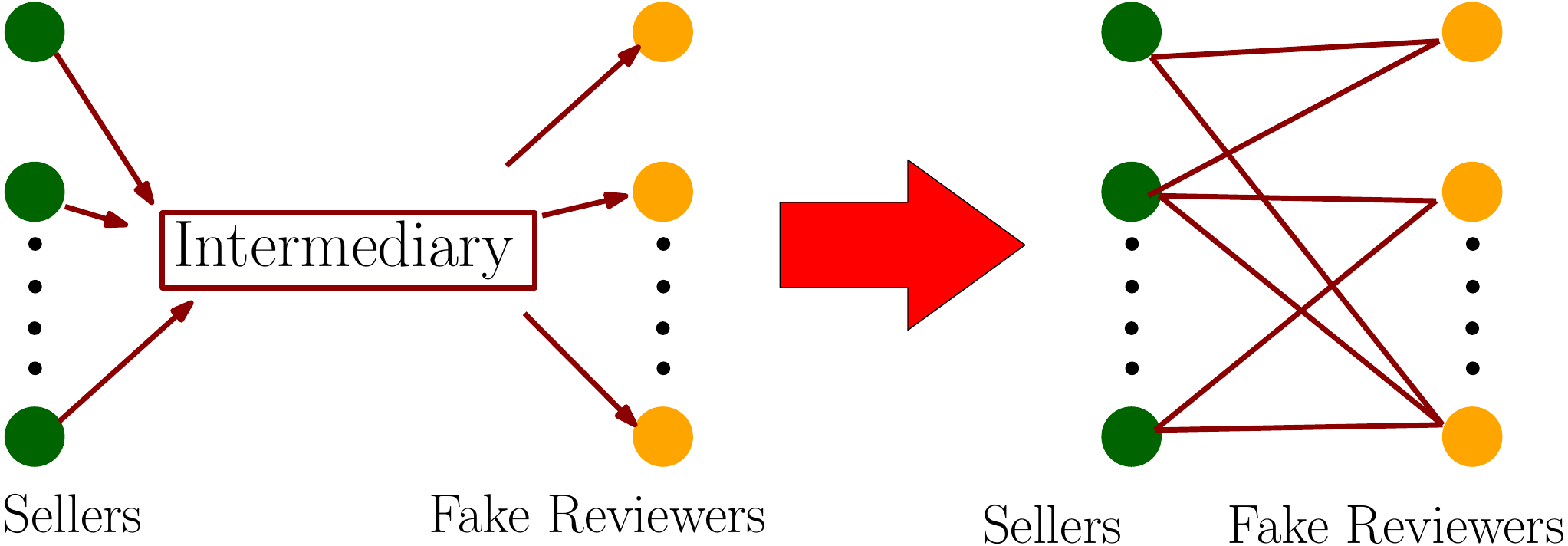}
\caption{\label{biPart:fig} Seller-Reviewer Data Anomaly: key signals.}
\end{figure}

\begin{figure}[t]
\center
\includegraphics[width=0.7\textwidth]{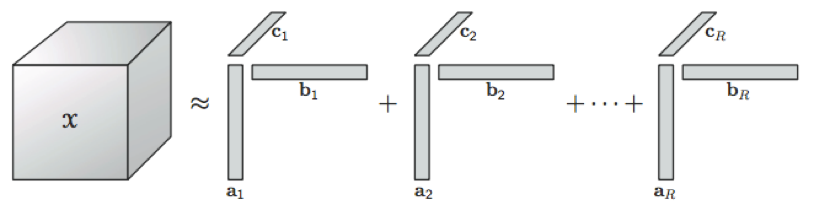}
\caption{\label{tensor:fig} CP tensor decomposition.}
\end{figure}

\subsubsection{CP Tensor Decomposition}

Let ${\mathcal Y}$ be a 3-mode tensor. We can decompose it as
\begin{equation} {\mathcal Y} = \sum_{r=1}^R \vec{a}_r \odot \vec{b}_r \odot \vec{c}_r, \label{tensorD:eqn} \end{equation}
where $\vec{a}_r$, $\vec{b}_r$, and $\vec{c}_r$ are vectors (or rank-1 tensors) and $\odot$ represents vector outer product. $R$ is called the rank of the tensor. This is called \textit{CP Decomposition} and is a generalization of matrix singular value decomposition. See Figure~\ref{tensor:fig}.

\subsection{Bayesian Tensor Factorization}
Bayesian tensor factorization based on Poisson, Negative-Binomial and Logistic formulation (described below) have been proposed in \citep{schein15a, hu15a, rai14a, rai15a, rai15b} to infer multi-mode relations, which can be also be applied to identify anomalies. All of these approaches are un-supervised except the Logistic CP decomposition model \citep{rai15b} that can also leverage features (via side-information) while factorizing the tensor.

\subsubsection{Bayesian Poisson Tensor Factorization} 

Equation~\ref{tensorD:eqn} represents a deterministic decomposition. For count tensors, the authors in \citep{schein15a} make it probabilistic by assuming a Poisson likelihood and call their model as Bayesian Poisson Tensor Factorization (BPTF). The generative model for BPTF is shown below:
\begin{eqnarray*}
      {\mathcal Y} & \sim & \text{Poisson}( \sum_{r=1}^R \vec{a}_r \odot \vec{b}_r \odot \vec{c}_r ) \\
      a_{r,i}, b_{r,i}, c_{r,i} & \sim & \text{Gamma}(\alpha, \alpha \beta^{(r)} ).
\end{eqnarray*}
The authors in \citep{schein15a} have implemented BPTF (un-supervised model) using batch algorithm and we have leveraged this as one of the baselines in our experiments. Given the Poisson model and that the partially labeled target information being binary or real, it does not seamlessly lend itself for semi-supervised extensions.

\subsubsection{Beta Negative-Binomial CP Model} 

Let ${\mathcal Y}$ be a $K$-mode tensor of size $n_1 \times n_2 \times \cdots \times n_{K}$. The authors in \citep{hu15a} assume a count tensor and hence define a Poisson likelihood. They call their model as Beta Negative-Binomial CP decomposition (BNBCP). BNBCP decomposition of tensor ${\mathcal Y}$ into $R$ components is as below:
\begin{equation*} 
{\mathcal Y} \sim \text{Poisson}(\sum_{r=1}^R \lambda_r \vec{u}^{(1)}_r \odot \cdots \odot \vec{u}^{(K)}_r),
\end{equation*}
where vectors $\vec{u}^{(k)}_r$ for $k \in [1,K]$ denote rank-1 tensors.
The generative model for BNBCP as shown below have Gamma and Dirichlet priors assigned to $\lambda_r$ and $\vec{u}^{(k)}_r$ respectively. The Poisson-Gamma hierarchical construction for $\mathcal{Y}$ effectively results in a \textbf{Negative Binomial} likelihood model which leads to better robustness against over-dispersed counts. Vectors \textit{$\vec{u}^{(k)}_r$} in the simplex can be seen as ``topics'' over the \textit{$n_{k}$} entities in mode \textit{k}. Rank \textit{R} of $\mathcal{Y}$ can be inferred from the gamma-beta construction on \textit{$\lambda_r$}'s:
\begin{eqnarray*}
\lambda_r \sim \text{Gamma}(g_r, \frac{p_r}{1-p_r}) \\
p_r \sim \text{Beta}(c\epsilon, (1-c)\epsilon) \\
\vec{u}^{(k)}_r \sim \text{Dirichlet}(a_{1}^{(k)}, \cdots, a_{n_{k}}^{(k)}).
\end{eqnarray*}
The BNBCP model is a fully conjugate model and inference is done using Variational Bayes (VB). To be able to scale for massive tensors, we have implemented online VB via Stochastic Variational Inference (SVI). We use this as one of the baselines in our experiments (un-supervised comparison only). This model can be seamlessly extended to achieve semi-supervised learning but the results were sub-par as compared with the results obtained from our semi-supervised Logistic CP model. Hence its semi-supervised results are not compared in this paper.

\subsubsection{Logistic CP Decomposition} 

In Logistic CP tensor factorization framework \citep{rai14a, rai15a, rai15b} the decomposition is as follows:
\begin{equation*} 
{\mathcal Y} \sim f(\sum_{r=1}^R \lambda_r \vec{u}^{(1)}_r \odot \cdots \odot \vec{u}^{(K)}_r), 
\end{equation*}
where $f$ specifies the Bernoulli-Logistic function for the binary valued tensor.
In mode $k$, consider entity $i_k$. Denote $\vec{u}^{(k)}_{i_k}$ as the $R$ dimensional factor corresponding to entity $i_k$.
\begin{enumerate}
\item Gaussian priors are assigned to latent variables $\vec{\lambda}$ and $\vec{u}^{(k)}_{i_k}$ for $k \in [1, K]$.
\item The number of non-zero values of $\vec{\lambda}$ determines the rank of the tensor. The variance of the Gaussian prior for $\vec{\lambda}$ is controlled by a \textit{Multiplicative Gamma Process} that has the property of reducing the variance as $r$ increases.
\item Given the logistic formulation for the tensor decomposition, to obtain closed form updates the data is augmented via additional variables $\omega$ that are P\'{o}lya-Gamma distributed.
\item We impose non-negativity constraint on $\vec{u}^{(k)}_{i_k}$ for all modes $k$.
\end{enumerate}
We use this un-supervised model as our base and have developed semi-supervised extensions to it as described in section \ref{sec:semi_supervised}. \citep{rai14a, rai15a, rai15b} propose using either sufficient statistics (in batch EM or as online EM) or Gibbs sampling (in batch) for inference. They claim that the online EM reaches reasonably good quality solutions fairly quickly as compared with their batch counterparts in most situations. However their online EM does scalar updates of each latent variable that is inherently a vector of dimension $R$ - which may result in slower convergence. We propose using partial natural gradient learning in a stochastic setting for inference that is both: online in nature as well as does vectorized updates for each of the latent variables.

Natural gradient learning requires computation of the \textit{Fisher} information matrix (as well as its inverse) that is obtained by taking the expectation of the \textit{Hessian} matrix w.r.t. the data. In most models, computing the Fisher information matrix is non-trivial and even if it can be computed, its inverse is usually not tractable. We show that the P\'{o}lya-Gamma data augmentation facilitates easy computation of the Fisher information matrix. To make the inverse of this matrix tractable, we exploit the quadratic structure of the loss function in each of the latent variables to be able to work with only the partial Fisher information matrix instead that is much smaller in size as compared with the full Fisher information matrix. Section \ref{sec:Mstep} describes this in greater detail.

We empirically show that in the semi-supervised setting, on the test data, scalar updates of vector parameters in Online-EM is sub-optimal i.e., results in lower ROC-AUC and precision as compared with the partial natural gradient algorithm. We also show that partial natural gradient algorithm produces better ROC-AUC and precision than stochastic gradient learning in the semi-supervised setting, specifically in detecting abusive sellers. So far, we have not come across any literature that have applied partial natural gradient learning for inference in Bayesian CP tensor decomposition.

\section{Semi-Supervised Logistic CP Decomposition}\label{sec:semi_supervised}
In this section we describe our semi-supervised extensions to the Logistic CP model for anomaly detection. We have prior information associated with a subset of the entities for at least one of the modes. This prior information is specified as a target (either binary or real) that corresponds to a specific type of abuse. Our framework is called ``semi-supervised tensor decomposition'' since the tensor decomposition is achieved by simultaneously incorporating the prior data i.e., the target information given for a subset of the entities of a mode(s). The intuition behind using the target information is that the patterns hidden in the known abusive entities could be leveraged to discover more entities that have similar signatures and with greater precision. Tensor decomposition with such target information is as follows:
\begin{enumerate}
\item Target information is specified for at least one of the modes.
\item Target information in each mode can either be real numbers or binary labels.
\begin{enumerate}
\item If data is binary; then both positive and negative labels need to be specified (positive labels indicate abuse and negative labels indicate no-abuse).
\item Data can be specified for only a subset of the entities in that mode (semi-supervised learning).
\item The factorization of the tensor is achieved by taking the target information across the mode(s) into account.
\end{enumerate}
\end{enumerate}

This paper concerns only with binary targets and all our experiments have been performed using binary targets. Binary target for mode $k$ is specified as a matrix with $M$ rows; where $M$ denotes the number of entities that have binary labels specified such that $M < n_k$ for semi-supervised learning. The first column of the matrix consists of the entity identifiers and the second column consists of corresponding binary labels.

Let $z_{n}^{(k)}$ denote the binary label (either $+1$ or $-1$) associated with entity $n$ in mode $k$. CP decomposition of a tensor produces $R$ rank-1 tensors of length $n_k$ for each mode $k$; which can also be viewed as a factor matrix of dimension $n_k \times R$ for mode $k$. Consider entity $n$ in mode $k$ that has a binary label associated with it. For entity $n$, denote its corresponding row in the factor matrix by $\vec{u}_{i_k=n}^{(k)}$ where $\vec{u}_{i_k=n}^{(k)}$ forms one instance of the explanatory variables. Note that $\vec{u}_{i_k=n}^{(k)}$ is a $R$ dimensional vector, where each element of this vector is denoted by $u_{i_k=n,r}^{(k)}$. Since $M$ entities in mode $k$ have binary labels associated with them, the design matrix consists of the corresponding $M$ rows of the factor matrix. Let $\vec{\hat{\beta}}^{(k)}$ denote the vector of $R$ coefficients with the bias denoted as $\beta_{0}^{(k)}$ for mode $k$ and let $\vec{\beta}^{(k)}$ denote the vector of $R+1$ coefficients that includes the bias.
The logistic formulation for semi-supervised learning is:
\begin{equation*} P(z_{n}^{(k)} = 1) = \text{Logistic}(\vec{\hat{\beta}}{^{(k)}}^{\top}\vec{u}_{i_k=n}^{(k)} + \beta_{0}^{(k)}). \end{equation*}
The coefficients ($\vec{\beta}^{(k)}$) for mode $k$ are assigned Gaussian priors. To get closed form updates of the coefficients and factors; we introduce auxiliary variables denoted by $\nu_{n}^{(k)}$ that are P\'{o}lya-Gamma distributed for each entity $n$ in mode $k$ that has a binary label associated with it. The next section provides a detailed description of the modeling.

Note that the binary target is usually specified (and available) for only a small subset of the entities of mode $k$. Hence the learning is semi-supervised. Based on the labels for a subset of entities; the tensor decomposition technique can infer the neighbors of these entities via information present in other modes. For example; if a seller $a$ is flagged as having review abuse; then the tensor decomposition technique would infer the reviewers associated with seller $a$ during some time $\Delta t$ where the density is high. This will facilitate detection of other sellers who are also connected to some subset of these same reviewers during the same time interval $\Delta t$. These other sellers would then have a high probability of being abusive.

We apply an online (stochastic) algorithm to infer the values of all the latent variables of the semi-supervised Logistic CP model. The latent variables being $\vec{\lambda}$ of length $R$, matrix $\boldsymbol{U}^{(k)}$ for each mode $k \in [1, K]$ of dimension $n_k \times R$ and $\vec{\beta}^{(k)}$ of length $R+1$ for each mode $k \in [1, K]$ that has target information. 

The next two sections describe the modeling and inference of the latent variables in a stochastic setting for the semi-supervised Logistic CP model.

\section{Model Description for Latent Variables}\label{sec:Estep}

The sub-sections below describe in detail the modeling for all the latent variables and provides the update equations for the hyper-parameters of the semi-supervised Logistic CP model.

\subsection{Model for $\vec{\lambda}$}
$\vec{\lambda}$ has a Gaussian prior whose variance is determined via Multiplicative Gamma Process \citep{bhattacharya11a, durante16a}. It falls under the \textit{Adaptive Dimensionality Reduction} technique which adaptively induces sparsity and automatically deletes redundant parameters not required to characterize the data. The Multiplicative Gamma Process consists of a multiplicative Gamma prior on the precision of a Gaussian distribution that induces a multiplicative Inverse-Gamma prior on its variance parameter. However its performance is sensitive to the hyper-parameter settings of the Inverse-Gamma prior and hence it is very important to not naively choose the hyper-parameter values. But instead follow certain strategies to set their values that are based on their probabilistic characteristics.
The generative model for $\lambda_r$ for $r$ in $[1, R]$ and the Multiplicative Gamma Process are:
\begin{eqnarray*}
\delta_l \sim \text{Inv-Gamma}(a_l, 1) \\ 
\lambda_r \sim \mathcal{N}(0, \tau_r),
\end{eqnarray*}
where:
\begin{equation} \label{tauEq:eqn}
\tau_r = \prod_{l=1}^{r} \delta_l.
\end{equation}
The idea is that for increasing \textit{r}, $\tau_r$ should be decreasing in a probabilistic sense. In other words, the following stochastic order should be maintained with high probability, i.e.,
\begin{equation*} 
\tau_1 \geq \tau_2 \geq \cdots \geq \tau_R. 
\end{equation*}
To guarantee such a stochastic order with a high probability, as suggested in \citep{durante16a}, we need to set $a_1 > 0$ and $a_r > a_{r-1}$ and non-decreasing for all $r > 1$. Hence we choose the hyper-parameters values as follows:
\begin{eqnarray*}
a_1 = 1 \\ 
a_r = a_1 + (r - 1).\frac{1}{R}.
\end{eqnarray*}
The update for $\delta_r$ in iteration \textit{t} is given by (see  \citep{rai14a} for details):
\begin{equation} 
\delta_r = \frac{1 + \sum_{h = r}^{R} \frac{\lambda_h^{2}}{2} \prod_{l = 1, l \neq r}^{h} \frac{1}{\delta_l}}{0.5(R - r + 1) + a_r + 1}. \label{deltaEq:eqn} 
\end{equation}
From (\ref{tauEq:eqn}), we can calculate the $\tau_r$s in iteration \textit{t} for $r \in [1,R]$. Denote $\vec{\tau}$ as the vector consisting of $\tau_r$ for $r \in [1,R]$.

Let $\lambda_r$ for $r \in [1,R]$ denote an element of $\vec{\lambda}$. We introduce auxiliary variables (P\'{o}lya-Gamma distributed variables \citep{polson, pillow1}), denoted by $\omega_i$ for each element $i$ of the input tensor data, via data augmentation technique.

Consider a mini-batch defined at iteration $t$ as $I_t$. Define for each $i \in I_t$; $\phi_i = \vec{\lambda}^{T}A_i$ where: $A_i$ denotes a vector consisting of elements $A_i^{r} = \prod_{k = 1}^{K}u_{i_k,r}^{(k)}$ for $r \in [1, R]$. Let $A$ be the matrix whose rows are $A_i$ for $i \in I_t$. Let $\hat{\omega}_i$ for $i \in I_t$ be the expected value of the auxiliary variable $\omega_i$ corresponding to the $i^{th}$ element of the input tensor data. The expected value of $\omega_i$ has a closed-form solution given by:
\begin{equation*} E[\omega_i] = \hat{\omega}_i = \frac{\tanh(\frac{\phi_{i}}{2})}{2\phi_{i}}. \end{equation*}
The update for $\vec{\lambda}$ in iteration \textit{t}, with the current mini-batch $I_t$, is obtained by maximizing the natural logarithm of the posterior distribution of $\vec{\lambda}$ given by:
{\small
\begin{equation}
\log[g(\vec{\lambda})] = \max_{\vec{\lambda}} \Bigl[ \Bigl[\sum_{i \in I_t} \kappa_i\phi_i - \frac{\omega_i\phi_i^{2}}{2}\Bigr] - \Bigl[\frac{(\vec{\lambda} \oslash \sqrt{\vec{\tau}})^\top(\vec{\lambda} \oslash \sqrt{\vec{\tau}})}{2}\Bigr] \Bigr],
%g(\vec{\lambda}) := \exp\Bigl[\sum_{i \in I_t} \kappa_i \phi_i - \frac{\hat{\omega_i} \phi_i^{2}}{2}\Bigr] \exp\Bigl[-\frac{(\vec{\lambda} \oslash \sqrt{\vec{\tau}})^\top (\vec{\lambda} \oslash \sqrt{\vec{\tau}})}{2}\Bigr],
\label{fMax:eqn}
\end{equation}
}where $\kappa_{i} = y_i - \frac{1}{2}$ and $y_i \in \{0, 1\}$. And the operator $\oslash$ represents element-wise division between the two vectors $\vec{\lambda}$ and $\vec{\tau}$.
Note that (\ref{fMax:eqn}) is a quadratic equation in $\phi_i$ i.e., $\vec{\lambda}$ and hence has a \emph{closed-form} update.

\subsection{Model for $\vec{\beta}^{(k)}$ for mode $k$}

The generative model for the coefficients $\beta_{r}^{(k)}$ for $r \in [0,R]$ is:
\begin{eqnarray*}
\beta_{r}^{(k)} \sim \mathcal{N}(0, \rho_{r}{^{(k)}}^{2}) \\
\rho_{r}{^{(k)}}^{2} \sim \text{Inv-Gamma}(a_c, b_1).
\end{eqnarray*}
The Inverse-Gamma hyper-prior on the variance parameter of the Gaussian distribution provides \textit{adaptive L2-Regularization}. The parameters of the Inverse-Gamma distribution ($a_c = 1$ and $b_1 = 0.6$) have been chosen so as to provide just the right amount of regularization for the coefficients $\vec{\beta}^{(k)}$.

Let $M$ denote the number of entities in mode $k$ that have binary labels. Let $\tilde{\vec{u}}_{i_k=m}$ denote $\vec{u}_{i_k=n}$ prepended with $1$, to account for the bias. The logistic function i.e., likelihood $\mathcal{L}_{m}^{(k)}$ corresponding to entity $m$ with label $z_{m}^{(k)} = 1$ is given by:
\begin{equation*} 
\mathcal{L}_{m}^{(k)} = \frac{1}{1 + exp[-z_{m}^{(k)}{\vec{\beta}{^{(k)}}^{\top}\tilde{\vec{u}}_{i_k=m}]}}. \end{equation*}
With introduction of P\'{o}lya-Gamma variables \citep{polson, pillow1} denoted by $\vec{\nu}^{(k)}$ for task $l$; the joint likelihood corresponding to entity $m$ that includes the data augmented variable $\nu^{(k)}_m$ becomes:
\begin{equation*} 
\mathcal{L}_{m}^{(k)} = exp(z_{m}^{(k)}\frac{\psi_{m}^{(k)}}{2} - \nu_{m}^{(k)}\frac{\psi_{m}{^{(k)}}^{2}}{2}), 
\end{equation*}
where:
\begin{equation*}
\begin{split}
\psi_{m}^{(k)} = \vec{\beta}{^{(k)}}^{\top}\tilde{\vec{u}}_{i_k=m} \ \ \text{and} \\
\mathbb{E}[\nu_{m}^{(k)}] = \hat{\nu}_{m}^{(k)} = \frac{\tanh(\frac{\psi_{m}^{(k)}}{2})}{2\psi_{m}^{(k)}}.
\end{split}
\label{nuPG:eqn}
\end{equation*}
The update for $\vec{\beta}^{(k)}$ in iteration \textit{t} is obtained by maximizing the natural logarithm of the posterior distribution of $\vec{\beta}^{(k)}$ given by:
\begin{equation}
\begin{split}
F(\vec{\beta}^{(k)})=\max_{\vec{\beta}^{(k)}}\Bigl[
\sum_{m = 1}^{M} [z_{m}^{(k)}\frac{\psi_{m}^{(k)}}{2} - \nu_{m}^{(k)}\frac{\psi_{m}{^{(k)}}^{2}}{2}]\\
- \frac{(\vec{\beta}^{(k)} \oslash \vec{\rho}^{(k)})^\top (\vec{\beta}^{(k)} \oslash \vec{\rho}^{(k)})}{2}]
\Bigr].
\end{split}
\label{fBeta:eqn}
\end{equation}
Equation (\ref{fBeta:eqn}) is a quadratic equation in $\vec{\beta}^{(k)}$ and hence has a \emph{closed-form} update. And the operator $\oslash$ represents element-wise division between the two vectors $\vec{\beta}^{(k)}$ and $\vec{\rho}^{(k)}$.

Subsequently, the update for $\rho_{r}{^{(k)}}^{2}$ at time step $t$ is given by:
\begin{equation}\label{eq:update_rho}
{\rho_{r}{^{(k)}}^{2}}_{(t)} = \frac{{\beta_{r}{^{(k)}}^{2}}_{(t-1)}}{2a_c + 3} + \frac{2b_1}{2a_c + 3}.
\end{equation}

\subsection{Model for $\vec{u}_{i_k,r}^{(k)}$ for mode $k$}

The generative model for the factors $u_{i_k,r}^{(k)}$ for $r \in [1,R]$ is:
\begin{equation*}
\begin{array}{l}
u_{i_k,r}^{(k)} \sim \mathcal{N}(0, \mu_{i_k, r}^{2})\\
\mu_{i_k,r}^{2} \sim \text{Inv-Gamma}(a_c, b_1)\qquad  \text{with target information} \\
\mu_{i_k,r}^{2} \sim \text{Inv-Gamma}(a_c, b_2) \qquad   \text{without target information}.
\end{array}
\end{equation*}
The Inverse-Gamma hyper-prior on the variance parameter of the Gaussian distribution provides \textit{adaptive L2-Regularization}. The Inverse-Gamma parameters are set so that greater amount of regularization is provided for the mode that has target information than the mode that does not have target information, hence we set $b_2 = 9$ that results in minimal regularization. The reason being that the factors corresponding to the mode with target information also act as co-variates (explanatory variables) in predicting the binary target. Note that (from the previous sub-section), we apply the same amount of regularization to the corresponding coefficients $\vec{\beta}^{(k)}$.

Denote $\vec{u}_{k,i_k}$ as the $R$ dimension vector consisting of factors $u_{i_k,r}^{(k)}$ corresponding to entity $i_k$ in mode $k$ and $r \in [1, R]$. Define for each $i \in I_t$ and $i_k = n$; $\phi_i = (\vec{u}_{i_k = n}^{(k)})^{T}\vec{C}_{i_k=n}$ where
each element of $\vec{C}_{i_k=n}$ is $C_{i_k=n,r} = \lambda_r.\prod_{k^{'} \neq k}^{K} u_{i_{k^{'}},r}^{(k^{'})}$ for $r \in [1, R]$. Let $\boldsymbol{C}^{(k)}$ be the matrix whose rows are $\vec{C}_{i_k}$.

\textbf{Mode $k$ without target information}:
The update for $\vec{u}_{i_k=n}^{(k)}$ in iteration \textit{t} is obtained by maximizing the natural logarithm of the posterior distribution of $\vec{u}_{i_k=n}^{(k)}$ given by:
{\small
\begin{equation}
\begin{array}{l}
F(\vec{u}_{i_k=n}^{(k)})=\max_{\vec{u}_{i_k=n}^{(k)}}\Bigl[ 
\sum_{i \in I_t: i_k = n} [\frac{\phi_i}{2} - \frac{\omega_i\phi_i^{2}}{2}]\\
 \qquad \ \- \frac{(\vec{u}_{i_k=n}^{(k)} \oslash \vec{\mu}_{i_k=n})^\top (\vec{u}_{i_k=n}^{(k)} \oslash \vec{\mu}_{i_k = n})}{2} + (\vec{\zeta}_{i_k=n})^{\top}\vec{u}_{i_k=n}^{(k)}
\Bigr],
\end{array}
\label{fUeq:eqn}
\end{equation}
}where $\vec{\zeta}_{i_k=n}$ is the vector of Lagrange multipliers for the non-negativity constraint on the $\vec{u}_{i_k=n}^{(k)}$. And the operator $\oslash$ represents element-wise division between the two vectors $\vec{u}_{i_k=n}^{(k)}$ and $\vec{\mu}_{i_k = n}$.

Equation (\ref{fUeq:eqn}) is a quadratic equation in $\vec{u}_{i_k=n}^{(k)}$ with non-negativity constraints and hence has a \emph{closed-form} update.

Subsequently the update for $\mu_{i_k=n,r}^{2}$ at time step $t$ is given by:
\begin{equation}\label{eq:update_mu_without_target}
{\mu_{i_k=n,r}^{2}}_{(t)} = \frac{{u_{i_k=n,r}{^{(k)}}^{2}}_{(t-1)}}{2a_c + 3} + \frac{2b_2}{2a_c + 3}.
\end{equation}

\textbf{Mode $k$ with binary target information}:
Given binary target information for mode $k$, let $z_{n}^{(k)}$ denote the binary label (either $+1$ or $-1$) for entity $n$ in mode $k$. Let $\nu_{n}^{(k)}$ correspond to the data augmented variable that is P\'{o}lya-Gamma distributed for the entity $n$.

The update for $\vec{u}_{i_k=n}^{(k)}$ in iteration \textit{t} is obtained by maximizing the natural logarithm of the posterior distribution of $\vec{u}_{i_k=n}^{(k)}$ given by:
\begin{equation}
\begin{array}{ll}
F(\vec{u}_{i_k=n}^{(k)})=\max_{\vec{u}_{i_k=n}^{(k)}}\Bigl[
\sum_{i \in I_t: i_k = n} [\frac{\phi_i}{2} - \frac{\omega_i\phi_i^{2}}{2}] \\
\qquad - \frac{(\vec{u}_{i_k=n}^{(k)} \oslash \vec{\mu}_{i_k=n})^\top (\vec{u}_{i_k=n}^{(k)} \oslash \vec{\mu}_{i_k = n})}{2}\\
\qquad + [z_{n}^{(k)}\frac{(\vec{u}_{i_k=n}^{(k)})^{\top}\vec{\hat{\beta}}^{(k)}}{2} - \frac{\nu_{n}^{(k)}[\beta_{0}^{(k)} + (\vec{u}_{i_k=n}^{(k)})^{\top}\vec{\hat{\beta}}^{(k)}]^{2}}{2}]
\Bigr].
\end{array}
\label{fUeqSI:eqn}
\end{equation}
Note that (\ref{fUeqSI:eqn}) does not have the non-negativity constraint since we are training a Logistic model using the binary target information to detect abusive entities. And the operator $\oslash$ represents element-wise division between the two vectors $\vec{u}_{i_k=n}^{(k)}$ and $\vec{\mu}_{i_k = n}$. Also (\ref{fUeqSI:eqn}) is a quadratic equation in $\vec{u}_{i_k=n}^{(k)}$ and hence has a \emph{closed-form} update.

Subsequently the update for $\mu_{i_k=n,r}^{2}$ at time step $t$ is given by:
\begin{equation}\label{eq:update_mu_with_target}
{\mu_{i_k=n,r}^{2}}_{(t)} = \frac{{u_{i_k=n,r}{^{(k)}}^{2}}_{(t-1)}}{2a_c + 3} + \frac{2b_1}{2a_c + 3}.
\end{equation}

\section{Partial Natural Gradients: Inference} \label{sec:Mstep}

Natural gradient learning (or in the context of online learning) is defined in \citep{amari98a}. Natural gradient learning is an optimization method that is traditionally motivated from the perspective of information geometry and works well for many applications as an alternate to stochastic gradient descent, see {\citep{martens14a}. Natural gradient descent is generally applicable to the optimizaton of probabilistic models that uses \textit{natural gradient} in place of the standard gradient. It has been shown that in many applications, natural gradient learning seem to require far fewer total iterations than gradient descent hence making it a potentially attractive alternate method. However it has been known that for models with very many parameters, computing the natural gradient is impractical since it requires computing the inverse of a large matrix i.e., the Fisher information matrix. This problem has been addressed in prior works via using one of various approximatons to the Fisher that are designed to be easier to compute, store and invert than the exact Fisher.

Given that our model also has a lot of parameters, we have addressed this problem by exploiting the problem structure that facilitates working with partial Fisher (i.e., computing partial natural gradients) as described subsequently. We apply partial natural gradient learning to update the values of the latent variables of the semi-supervised Logistic CP model. Partial natural gradients implies that we only work with diagonal blocks of the Fisher information matrix instead of the full Fisher information matrix. Note that the latent variables in the semi-supervised model that we need to infer are $\vec{\lambda}$ of length $R$, matrix $\boldsymbol{U}^{(k)}$ for each mode $k \in [1, K]$ of dimension $n_k \times R$ and $\vec{\beta}^{(k)}$ of length $R+1$ for each mode $k \in [1, K]$ that has target information.

%\subsection*{Challenges in Applying Natural Gradient\\ Learning}
Natural gradient update in iteration $t$ is defined as:
{\small
\begin{equation}
\begin{array}{ll}
%\vec{\lambda}_{(t)} = \vec{\lambda}_{(t-1)} + \gamma_t.\mathcal{I}(.)^{-1}.\mathop{\mathbb{E}}_{\vec{y}, \omega_i:i \in I_t \boldsymbol{U}^{(k)}: k \in [1,K]} \Bigl[\nabla_{\vec{\lambda}} \log[g(\vec{\lambda})]\Bigr] \\
%\vec{\beta}^{(k)}_{(t)} = \vec{\beta}^{(k)}_{(t-1)} + \gamma_t.\mathcal{I}(.)^{-1}.\mathop{\mathbb{E}}_{\vec{z}, \nu_m^{(k)}:m \in [1,M], \boldsymbol{U}^{(k)}} \Bigl[\nabla_{\vec{\beta}}^{(k)} F(\vec{\beta}^{(k)})\Bigr] \\
%\vec{u}_{i_k = n,(t)}^{(k)} = \vec{u}_{i_k=n,(t-1)}^{(k)} + \gamma_t.\mathcal{I}(.)^{-1}.\mathop{\mathbb{E}}_{\vec{y}, \vec{z}, \nu_{m=n}^{(k)}, \boldsymbol{U}^{(k)}} \Bigl[\nabla_{\vec{u}_{i_k=n}^{(k)}} F(\vec{u}_{i_k=n}^{(k)})\Bigr],
\vec{\lambda}_{(t)} = \vec{\lambda}_{(t-1)} + \gamma_t.\mathcal{I}(.)^{-1}.\mathop{\mathbb{E}}\Bigl[\nabla_{\vec{\lambda}} \log[g(\vec{\lambda})]\Bigr] \\
\vec{\beta}^{(k)}_{(t)} = \vec{\beta}^{(k)}_{(t-1)} + \gamma_t.\mathcal{I}(.)^{-1}.\mathop{\mathbb{E}}\Bigl[\nabla_{\vec{\beta}}^{(k)} F(\vec{\beta}^{(k)})\Bigr] \\
\vec{u}_{i_k = n,(t)}^{(k)} = \vec{u}_{i_k=n,(t-1)}^{(k)} + \gamma_t.\mathcal{I}(.)^{-1}.\mathop{\mathbb{E}}\Bigl[\nabla_{\vec{u}_{i_k=n}^{(k)}} F(\vec{u}_{i_k=n}^{(k)})\Bigr],
\end{array}
\label{sgd:eqn}
\end{equation}
} where $\gamma_t$ in (\ref{sgd:eqn}) is given by:
\begin{equation*}
%\gamma_t = \max\Bigl[\frac{1}{(\tau_{p} + t)^{\theta}}, \frac{Q}{\eta_{0} + Wt}\Bigr].
\gamma_t = \frac{1}{(\tau_{p} + t)^{\theta}}.
\end{equation*}
Where $\mathcal{I}(.)^{-1}$ in (\ref{sgd:eqn}) indicates the inversion of the Fisher information matrix (square matrix) in each iteration whose size could possibly be in the tens of thousands or more depending on the data. This impacts scalability i.e., could result in very expensive computations and might also pose numerical stability issues leading to an intractable inverse computation.

\begin{figure}[t]
\center
\includegraphics[scale=0.5]{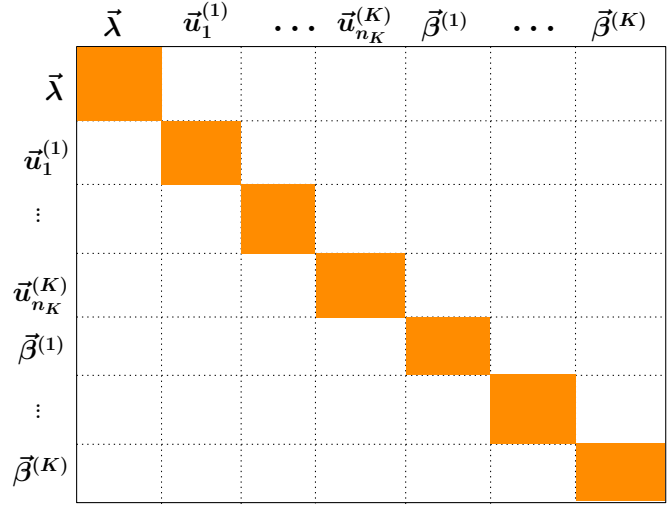}
\caption{\label{PNG2:fig} The block diagonal terms in the Fisher information matrix are strictly positive definite and are computationally easy to invert.}
\end{figure}

To circumvent this; we exploit the problem structure by noting the following:
\begin{enumerate}
\item Loss function is quadratic in each of the arguments ($\vec{\lambda}, \vec{u}^{(1)}_{1}, \allowbreak \ldots, \vec{u}^{(K)}_{n_K}, \vec{\beta}^{(1)}, \ldots, \vec{\beta}^{(K)}$).
\item This leads to a simpler approximation of the Fisher information matrix i.e., it facilitates working with the partial blocks (i.e., diagonal blocks) of the Fisher information matrix as shown in Figure~\ref{PNG2:fig}.
\end{enumerate}
Due to the \emph{individually} quadratic nature of the loss functions, each diagonal block is a symmetric positive definite matrix of size $R \times R$ for $\vec{\lambda}$ and $\vec{u}^{(k)}_{n_k}$ or $(R+1) \times (R+1)$ for $\vec{\beta}^{(k)}$. Hence the basic convergence guarantees for the full natural gradient learning extends to the partial set up as well; see \citep{bottou16a}. We note that computations of the partial Fisher information matrix is theoretically and numerically tractable as we are dealing with square matrices of size $R$ or ($R+1$), which is very small (value less than $10$) in our problem space.

Refer to the supplementary material for the detailed derivations of the partial Fisher information matrices as well as the gradients for each of the arguments, namely, $\vec{\lambda}$, $\vec{u}^{(1)}_{1} \ldots \vec{u}^{(K)}_{n_K}$ and $\vec{\beta}^{(1)} \cdots \vec{\beta}^{(K)}$.

The partial natural gradients are obtained when the corresponding gradient is multiplied by the inverse of the corresponding partial Fisher information matrix.
%\subsection*{Partial Natural Gradient Algorithm}
Algorithm~\ref{alg:naturalGrads} presents the pseudo-code for the semi-supervised CP tensor decomposition using natural gradients.

\section{Experiments}\label{sec:experiments}
For all our experiments, we have chosen the following values for the learning rate parameters $\tau_{p} = 256$ and $\theta = 0.61$.
%$Q = 0.024$, $\eta_0 = 6$ and $W = \frac{1}{\mathcal{N}_b}$ where $\mathcal{N}_b$ denotes the number of iterations in an epoch (depends on the mini batch-size).
We have chosen the following values for the parameters of the \textit{Inverse Gamma} distributions: $a_c = 1$, $b_1 = 0.6$ and $b_2 = 9$. We have set the mini-batch size to $512$ in all our simulations. These values have been chosen by performing 5-fold crossvalidations on a validation set. 

\subsection*{Dataset}
The binary tensor i.e., the input data corresponds to a random sampling of the products in the Amazon review data until October 2017. The modes of the tensor are reviewer ID, product ID, seller ID, rating and time. Note that rating corresponds to an integer between $1$ to $5$. Time is converted to a week index. The tensor consists of millions of entires, where each entry of the tensor represents a unique association that corresponds to a reviewer (buyer) $b$ rating a product $p$ from seller $s$ with a rating $r$ at time $t$.
%Note that in this experiment we are only using a single target i.e., single type of abuse though our semi-supervised approach can simultaneously handle multiple targets.

\subsection{Detection of Abusive Sellers}
We have partial ground truth data consisting of a small number of sellers; who have been actioned against for being guilty of review abuse - treated as positively labeled samples. To this we have included sellers who currently are not flagged with any kind of abuse - treated as negatively labeled samples. The negatively labeled sellers are roughly three times the number of positively labeled sellers. This forms the training set.
We have set aside an additional set of around $1500$ sellers as test set to measure the performance of the un-supervised and semi-supervised tensor decomposition techniques. The test set has similar distribution of positive and negative samples; where the abusive sellers (positively labeled set) have been identified in November and December of 2017 i.e., identified beyond the training time period.

\RestyleAlgo{boxruled}
\begin{algorithm}[H]
    \caption{Partial Natural Gradient.}
    \label{alg:naturalGrads}
    %\small
    \begin{enumerate}[1:]
\item Randomly initialize $\vec{\tau}, \vec{\lambda}$, $\vec{u}^{(1)}_{1} \ldots \vec{u}^{(K)}_{n_K}$ and $\vec{\beta}^{(1)} \cdots \vec{\beta}^{(K)}$.
\item Set the step-size schedule $\gamma_t$ appropriately.
\item \textbf{repeat}
\begin{enumerate}[a:]
\item Sample (with replacement) mini-batch $I_t$ from the training data.
%\item \textbf{repeat}
\item For $i \in I_t$ set \\
--\ \ \ \ \ \  $A_i^{r} = \prod_{k=1}^K u_{i_k,r}^{(k)}$ for $r \in [1,R]$ \\
-- \ \ \ \ \ \ $\hat{\omega}_i = \frac{\text{tanh}(\frac{\phi_i}{2})}{2\phi_i}$ where $\phi_i = \vec{\lambda}^{\top}A_i$ \\
-- \ \ \ \ \ \ $N_{ii} = \frac{1}{\Bigl[\text{exp}[-\frac{\phi_i}{2}] + \text{exp}[\frac{\phi_i}{2}] \Bigr]^2}$.
\item For $k \in [1, K]$ set \\
-- \ \ \ \ \ \ $C_{i_k=n,r}^{(k)} = \lambda_r\prod_{k' \ne k}^K u_{i_{k'},r}^{(k')}$ for $r \in [1,R]$ \\
-- \ \ \ \ \ \ For entity $i_k=m$ with binary target information $z_m^{(k)}$:\\
\ \ \ \ \ \ \ \ \ \ $\hat{\nu}_m^{(k)} = \frac{\text{tanh}(\frac{\psi_m^{(k)}}{2})}{2\psi_m^{(k)}}$ where $\psi_m^{(k)} = \vec{\beta}{^{(k)}}^{\top}\tilde{\vec{u}}_{i_k=m}$\\
\ \ \ \ \ \ \ \ \ \ \ $N_{m=n} = \frac{1}{\Bigl[\text{exp}[-\frac{\psi_{m=n}}{2}] + \text{exp}[\frac{\psi_{m=n}}{2}] \Bigr]^2}$ \\
-- \ \ \ \ \ \ Compute \textit{gradient} and \textit{partial Fisher information matrix}\\
\ \ \ \ \ \ \ \ \ \ \ w.r.t. $\vec{\beta}^{(k)}$ and update $\vec{\beta}^{(k)}$ using (\ref{sgd:eqn}). \\
-- \ \ \ \ \ \ Update $\rho_{r}{^{(k)}}^2$ using (\ref{eq:update_rho}).\\
-- \ \ \ \ \ \ Compute \textit{gradient} and \textit{partial Fisher information matrix}\\
\ \ \ \ \ \ \ \ \  w.r.t. $\vec{u}_{i_k:i \in I_t}^{(k)}$ and update $\vec{u}_{i_k:i \in I_t}^{(k)}$ using (\ref{sgd:eqn}). \\
-- \ \ \ \ \ \ Update $\mu_{i_k:i \in I_t,r}^2$ using (\ref{eq:update_mu_without_target}) or (\ref{eq:update_mu_with_target}).
\item  Compute \textit{gradient} and \textit{partial Fisher information matrix} w.r.t. $\vec{\lambda}$ and update $\vec{\lambda}$ using (\ref{sgd:eqn}).
\item Update $\vec{\tau}$ using (\ref{tauEq:eqn}).
\end{enumerate}
\item \textbf{until} forever
\end{enumerate}
\end{algorithm}

Table~\ref{tab:unsem} compares the relative performance of our semi-supervised approach with un-supervised approaches namely, BNBCP \citep{hu15a}, BPTF \citep{schein15a}, and M-Zoom \citep{shin16a} on the Precision, Recall and ROC-AUC measured on the test set. Among the un-supervised techniques, M-Zoom and BPTF are batch algorithms while BNBCP uses stochastic variational inference (our implementation). All the three flavors of our semi-supervised Logistic CP model implementations are stochastic in nature. Our proposed natural gradient based implementation is the only one that requires inverting a matrix in each iteration, hence it is approximately $2$\% slower than the online EM with sufficient statistics based implementation - which is the fastest among the three. The stochastic gradient based implementation is approximately $1$\% slower than the online EM with sufficient statistics based implementation.

For any column, the metric with the highest value is set at $100.0$. All other values are computed relative to this highest value in percent. Precision and Recall are measured on the test set taking the top $N$ sellers computed from each method; where $N$ corresponds to the number of unique sellers associated with the top $2$ blocks of M-Zoom output. ROC-AUC is computed on the test set. M-Zoom does not assign a score for each entity across the tensor modes but rather produces anomalous sub-tensors (i.e., blocks) as output. Hence ROC-AUC cannot be easily computed for M-Zoom.

\begin{table}[t]  
  \caption{Abusive sellers: un-Supervised and semi-supervised results - \textit{relative performance}.}
  \label{tab:unsem}
  \resizebox{\columnwidth}{!}{%
   \tiny{
  \begin{tabular}{ccccl}
    \toprule
    & \textbf{Method} & \textbf{Precision} & \textbf{Recall} & \textbf{AUC}\\
    \midrule
    \multirow{4}{*}{Un-Supervised}
    %& M-Zoom~\citep{shin16a} & 6.0 & \textbf{65} & -\\
    & M-Zoom~\citep{shin16a} & 83.8 & 83.3 & -\\
    %& BPTF~\citep{schein15a} & 7.0 & 69 & 59.8\
    & BPTF~\citep{schein15a} & 97.2 & 88.5 & 90.0\\
    %& BNBCP~\citep{hu15a} & 6.7 & 61 &  52.3\\
    & BNBCP~\citep{hu15a} & 93.3 & 78.2 &  78.8\\
    %& Logistic CP [Natural Gradient] & 6.8 & 0.171 & 51.5\
    & Logistic CP [\textit{Natural Gradient}] & 94.1 & 68.7 & 77.6\\
    \midrule
    \multirow{2}{*}{Semi-Supervised} 
    %& Logistic CP [Sufficient Statistics] & \textbf{6.0} & \textbf{72} & \textbf{60.4}\\
    & Logistic CP [\textit{Sufficient Statistics}] & 83.3 & 92.3& 91.0\\
    %& Logistic CP [Stochastic Gradient] & 6.4 & 69 & \textbf{61.2}\\
    & Logistic CP [\textit{Stochastic Gradient}] & 88.9 & 94.9 & 92.2\\
    %& Logistic CP [Natural Gradient] & \textbf{7.2} & 78 & \textbf{66.4}\\
    & Logistic CP [\textit{Natural Gradient}] & \textbf{100.0} & \textbf{100.0} & \textbf{100.0}\\
  \bottomrule
\end{tabular}
}
}
\end{table}

All the three flavors of our semi-supervised Logistic CP implementations have higher recall and AUC as compared with the un-supervised techniques. The best performing un-supervised method for AUC is BPTF and its value is $10$\% lower than our semi-supervised approach that uses partial natural gradient for inference. And the best performing un-supervised method for precision is also BPTF and its value is around $2.8$\% lower than our semi-supervised approach that uses partial natural gradient. The precision, recall and AUC for the semi-supervised approach that uses Online-EM with sufficient statistics (scalar updates) is lower as compared with the other two semi-supervised approaches indicating that scalar updates of vector parameters results in sub-optimal solutions.
%We think increasing the training set size as well as leveraging the multi-target framework should increase the performance gap between our semi-supervised approach and un-supervised approaches.

\subsection{Detection of Abusive Reviewers}
We have partial ground truth data of roughly $10$\% of reviewers who have been actioned against for being guilty of paid reviewer abuse - treated as positively labeled samples. To this we have included an almost equal number of reviewers who had the lowest scores from the un-supervised tensor decomposition model - treated as negatively labeled samples. This forms the training set.
We have an additional set of reviewers (roughly 4200), that is treated as our test set. In the test set; roughly $50$\% of the reviewers are labeled as positive i.e., identified as being guilty of paid reviewer abuse in the months of November and December 2017 to which we have added an almost equal number of reviewers that are labeled as negative.

Table~\ref{tab:unsem_rvr} compares the relative performance of our semi-supervised approach with un-supervised approaches namely, BNBCP \citep{hu15a}, BPTF \citep{schein15a}, and M-Zoom \citep{shin16a} on the metrics, namely, Precision, Recall and ROC-AUC. Precision and Recall are measured on the test set taking the top $N$ reviewers computed from each method; where $N$ corresponds to the number of unique reviewers associated with the top $2$ blocks of M-Zoom output. ROC-AUC is computed on the test set for all techniques except M-Zoom for the reason stated in the previous subsection.

\begin{table}[t]
  \caption{Abusive reviewers: un-Supervised and semi-supervised results - \textit{relative performance}.}
  \label{tab:unsem_rvr}
  \resizebox{\columnwidth}{!}{%
  \tiny{
  \begin{tabular}{ccccl}
    \toprule
    & \textbf{Method} & \textbf{Precision} & \textbf{Recall} & \textbf{AUC}\\
    \midrule
    \multirow{4}{*}{Un-Supervised}
    %& M-Zoom~\citep{shin16a} & 0.899 & 0.621 & -\\
    & M-Zoom~\citep{shin16a} & 97.2 & 91.3 & -\\
    %& BPTF~\citep{schein15a} & 0.922 & 0.665 & 0.505\\
    & BPTF~\citep{schein15a} & 99.7 & 97.8 & 96.7\\
    %& BNBCP~\citep{hu15a} & 0.917 & 0.592 & 0.468\\
    & BNBCP~\citep{hu15a} & 99.1 & 87.1 & 89.6\\
    %& Logistic CP [Natural Gradient] & 0.901 & 0.596 & 0.474\\
    & Logistic CP [\textit{Natural Gradient}] & 97.4 & 87.6 & 90.8\\
    \midrule
    \multirow{2}{*}{Semi-Supervised}
    %& Logistic CP [Sufficient Statistics] & \textbf{0.919} & \textbf{0.539} & \textbf{0.502}\\
    & Logistic CP [\textit{Sufficient Statistics}] & 99.4 & 79.3& 96.2\\
    %& Logistic CP [Stochastic Gradient] & 0.925 & 0.680 & 0.518\\
    & Logistic CP [\textit{Stochastic Gradient}] & \textbf{100.0} & \textbf{100.0} & 99.2\\
    %& Logistic CP [Natural Gradient] & \textbf{0.921} & \textbf{0.670} & \textbf{0.522}\\
    & Logistic CP [\textit{Natural Gradient}] & 99.6 & 98.5 & \textbf{100.0}\\
  \bottomrule
\end{tabular}
}
}
\end{table}

Our semi-supervised method with stochastic gradient and/or partial natural gradient learning have higher recall and AUC numbers as compared with all the un-supervised techniques. However the gain realized here is not as significant as was obtained in detecting abusive sellers when we look at the best performing un-supervised methods, namely BPTF and M-Zoom. The reason being that the seller behavior for a given product in the time period related to review abuse is more homogenous i.e., the product ratings that are obtained during the abusive period from some common pool of reviewers is statistically higher than normal. On the other hand, a reviewer may be colluding with more than one seller for providing fake reviews and at the same time may be a normal buyer for a product from a different set of sellers. This makes the training data for a reviewer more noisier and is hence providing lesser information for the semi-supervised methodology - resulting in smaller gains in precision, recall and AUC. There is very little difference in the performance between stochastic gradient and natural gradient, in fact stochastic gradient shows slightly better precision and recall numbers than obtained from natural gradient method. The recall numbers for the semi-supervised approach that uses Online-EM with sufficient statistics (scalar updates) is lower as compared with all the un-supervised approaches indicating that scalar updates of vector parameters results in sub-optimal solutions.

\subsection{Multi-mode Binary Target Information}
Our framework supports providing binary target information to more than one mode of the tensor simultaneously. Hence we also experimented with simultaneously providing the binary target data for both the seller and reviewer mode. However since the binary target information for the reviewer mode is noisy, the overall performance in detecting new sellers and reviewers were almost identical to the corresponding cases where we provide the binary target data to only one of the respective modes.

%\subsection{Partial Natural Gradient versus Baselines: Sufficient Statistics (Online-EM) \& Stochastic Gradient}
\subsection{Partial Natural Gradient versus Baselines}
We apply our semi-supervised Logistic CP tensor decomposition approach on the Amazon review data (5-mode tensor) to compare the performance of two baseline learning methods, namely, sufficient statistics (Online-EM) and stochastic gradient with our proposed partial natural gradient learning. Figure~\ref{ngdSgd:fig} shows the AUC plot for detecting abusive sellers and abusive reviewers across six epochs. Color red (solid) corresponds to natural gradient learning, color blue (dash-dot) corresponds to stochastic gradient learning and color black (dash) corresponds to online EM with sufficient statistics. We make the following observations:
\begin{enumerate}
\item Stochastic gradient learning has a tendency to over-train. The reason being that it is unable to shrink some of the values of $\lambda_r$ towards zero since that the tensor rank is less than $R$.
\item Partial natural gradient learning does not suffer from significant over-training since it is able to shrink 60\% of the values of $\lambda_r$ towards zero with-in the first one thousand iterations. This leads to similar AUC on train and test data sets for detecting abusive sellers as compared with the baselines. For detecting abusive reviewers however, the anomalous associations between the entities of the tensor mode in the test data is not necessarily indicative of abuse; hence the train and test AUC for all the three learning methods are much further apart.
\item For detecting abusive reviewers, at the end of six epochs; the test AUC is almost identical for both partial natural gradient and stochastic gradient learning. However for detecting abusive sellers, the test AUC is almost 8\% lower with stochastic gradient learning as compared with partial natural gradient learning.
\item For detecting abusive sellers, partial natural gradient shows better learning on test data than the baselines. For detecting abusive reviewers, though the AUC at the end of six epochs is similar for partial natural and stochastic gradient approaches, the AUC curve for partial natural gradient learning is much smoother as compared with the AUC curve from stochastic gradient learning.
\item Online-EM with sufficient statistics show poorer performance on test data (for both reviewers and sellers) when compared with performance of stochastic gradient or partial natural gradient learning.
\end{enumerate}

%\begin{figure}[t]
%\includegraphics[width=0.42\textwidth]{ngd_sgd_a.PNG}
%\caption{\label{ngdSgd:fig} ROC-AUC for partial natural gradient (red), stochastic gradient (black) \& sufficient statistics (brown).}
%\end{figure}

\begin{figure}[t]
\centering
\begin{tabular}{cc}
\noindent \begin{minipage}[b]{0.5\hsize}
\centering
\includegraphics[width=1.0\textwidth]{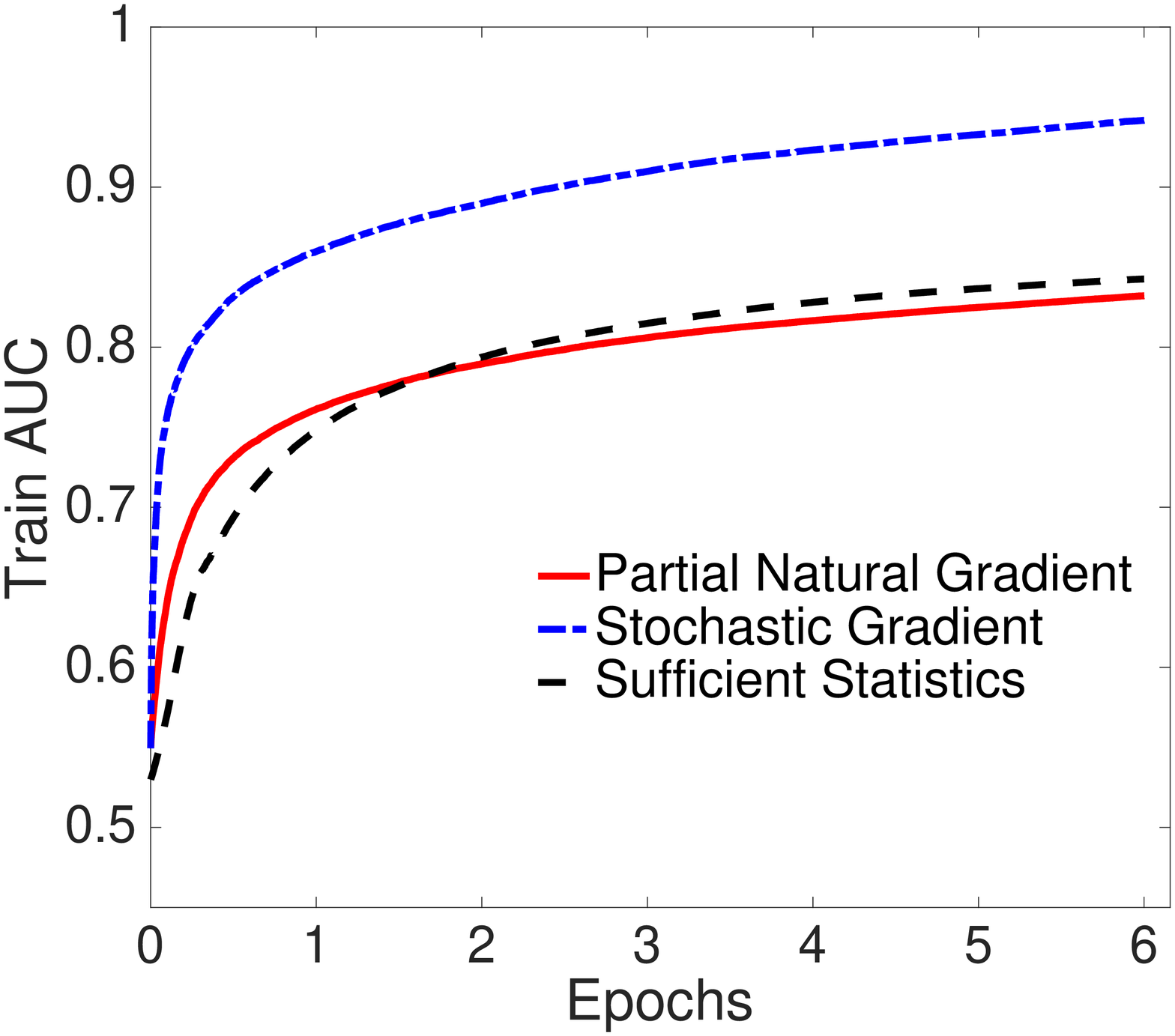}\\
{(a) Train data: sellers.}
\end{minipage}
\begin{minipage}[b]{0.5\hsize}
\centering
\includegraphics[width=1.0\textwidth]{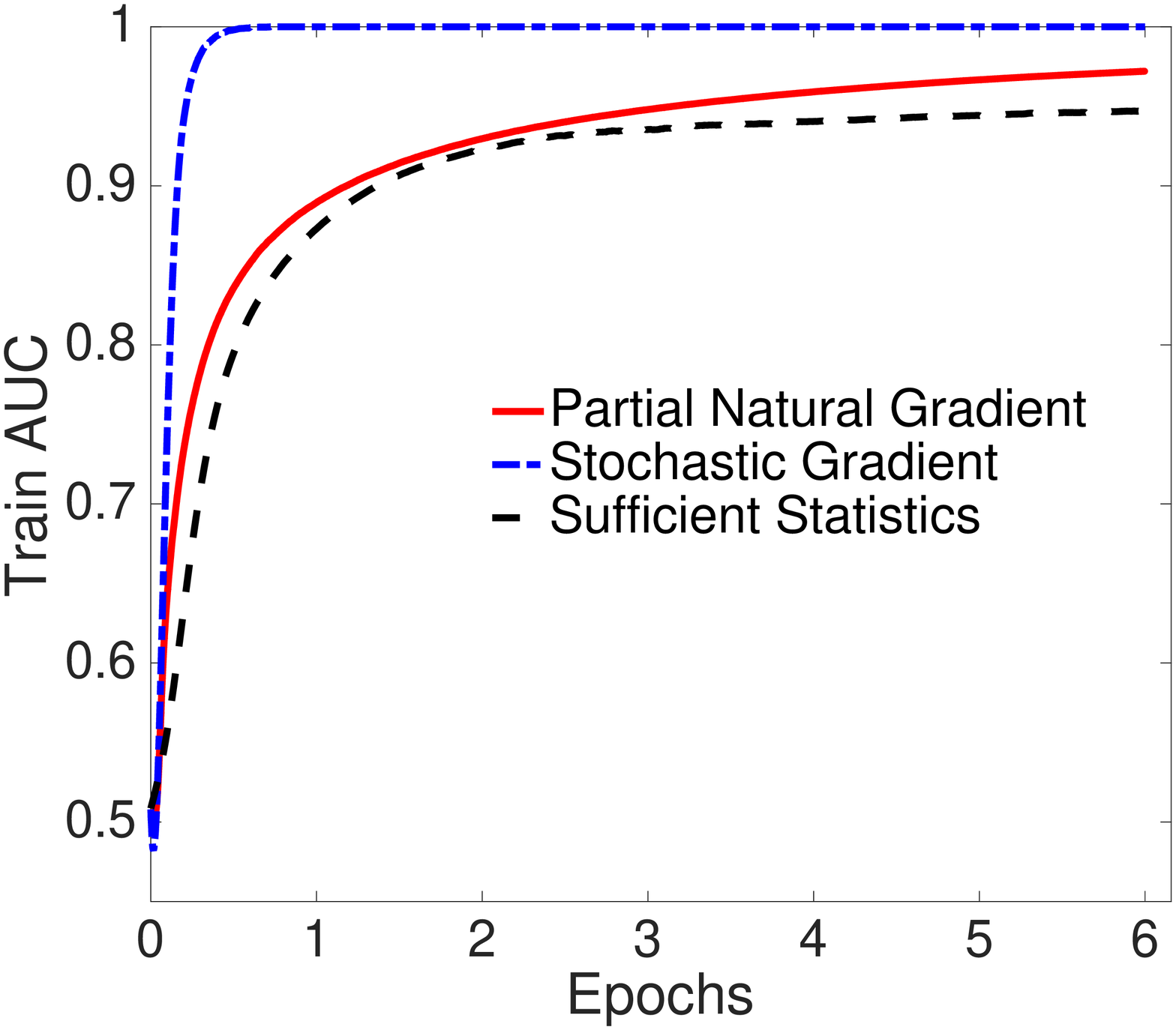}\\
{(b)  Train data: reviewers.}
\end{minipage}\\
\noindent \begin{minipage}[b]{0.5\hsize}
\centering
\includegraphics[width=1.0\textwidth]{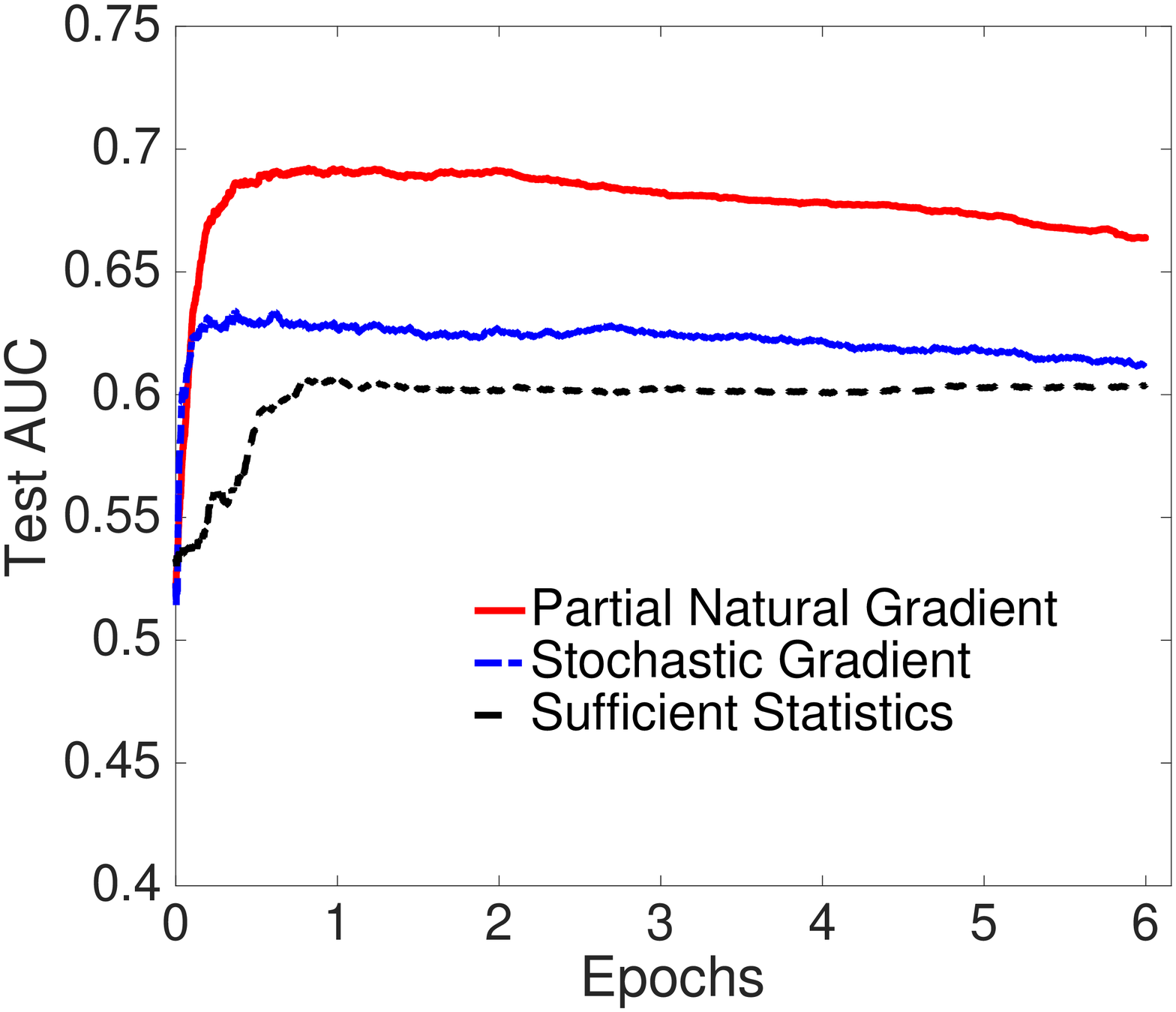}\\
{(c) Test data: sellers.}
\end{minipage}
\begin{minipage}[b]{0.5\hsize}
\centering
\includegraphics[width=1.0\textwidth]{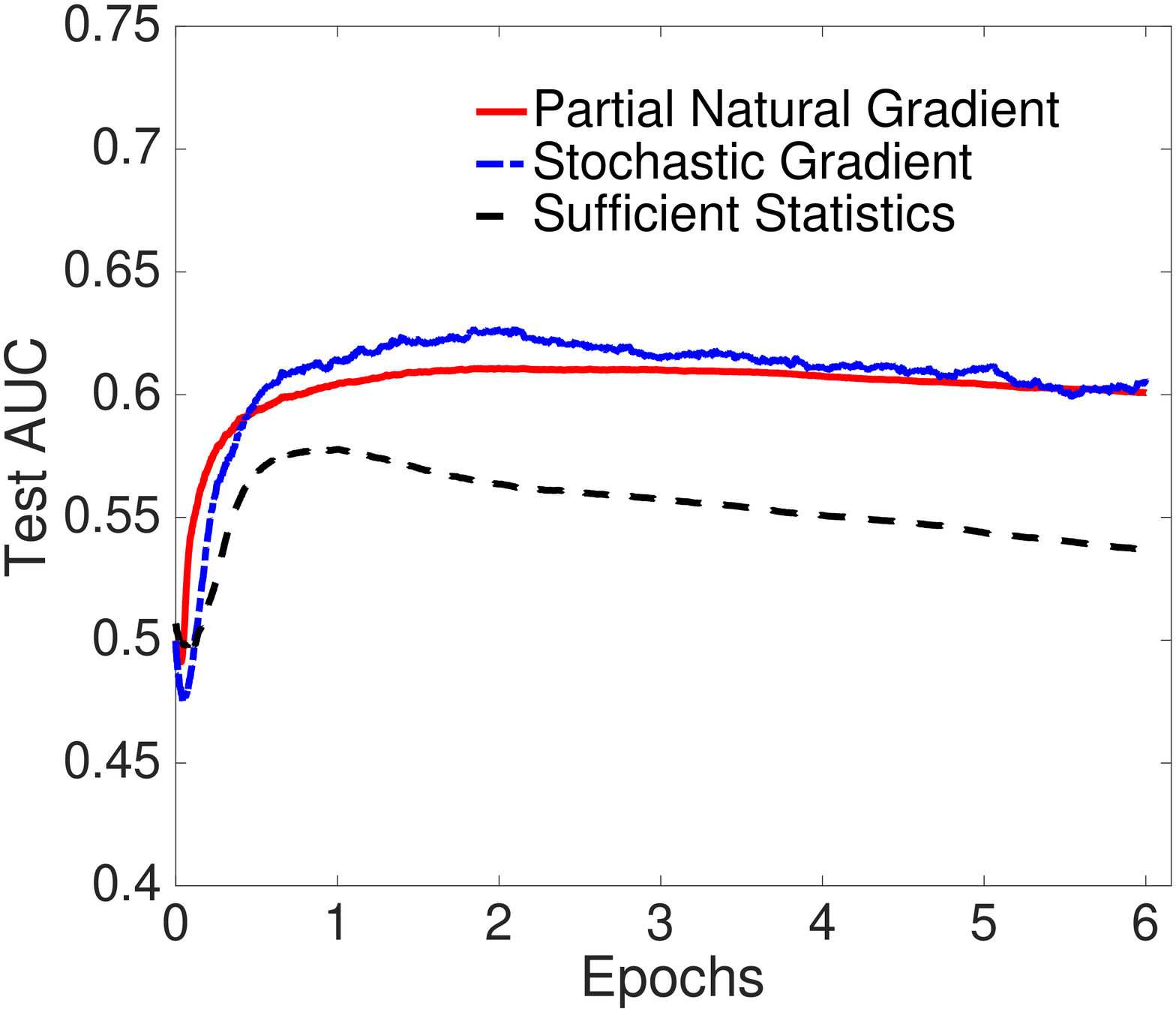}\\
{(d)  Test data: reviewers.}
\end{minipage}
\end{tabular}
\caption{Efficiency of partial natural gradient learning in identifying abusive sellers \& reviewers.}
\label{ngdSgd:fig}
\end{figure}

\section{Conclusions and Future Work}\label{sec:conclusions}
%\section{Conclusions}\label{sec:conclusions}
We have formulated anomaly detection as a semi-supervised binary tensor decomposition problem that can simultaneously incorporate binary target information for a subset of sellers and/or reviewers - based on the Logistic Model with P\'{o}lya-Gamma data augmentation. We have proposed natural gradient learning for inference of all the latent variables of the semi-supervised model and shown that the P\'{o}lya-Gamma formulation simplifies calculation of the partial Fisher information matrix. Our results have demonstrated that the proposed semi-supervised approach beats the state of the art unsupervised baselines and that our inference using partial natural gradient learning has shown better learning than online EM (using sufficient statistics) or stochastic gradient learning on test data sets from the time period that is non-overlapping with the training time period.

\textbf{Future work:} Our framework can be easily extended to do multi-target learning for each mode(s). For example, instead of considering only one form of abuse for the seller mode; in the multi-target domain we can simultaneously incorporate label information for different types of abuses. In such a setting, each binary target in a given mode corresponds to one type of abuse. We hypothesize that this would increase the precision/AUC of predicting new abusive entities since we can borrow information via accounting for the correlation across multiple forms of abuse. Such a multi-target learning could also be applied towards improving the performance of a recommender system. For example in the movies recommendation domain, we have the publicly available MovieLens data set. The MovieLens data consists of associations (hence can be considered as a binary tensor), where each association is a tuple of person, movie, time and rating. In this setting, we could incorporate gender, age-band and occupation type of a movie goer as multiple binary targets for the person mode. We could incorporate the genre of the movie as multiple binary targets for the movie mode. We hypothesize that such a semi-supervised CP tensor decomposition could result in recommending better movie choices that the movie goer might be interested to watch in the near future.

%\section{Acknowledgements}\label{sec:acks}
%We would like to thank Andrea Effgen, Michael Smyth and Kim Wilber from Amazon for their efforts in validating the sellers and reviewers from the model for abuse.

%% file: files/supp_arXiv-body.tex
\section{Computation of Partial Fisher information matrix} \label{supp_sec:fisher_matrix}

\subsection*{Partial Fisher information matrix w.r.t. $\vec{\lambda}$}

Consider the exponent of the function to be maximized w.r.t. $\vec{\lambda}$, which is:
{\small
\begin{equation}
g(\vec{\lambda}) := \exp\Bigl[\sum_{i \in I_t} \kappa_i \phi_i - \frac{\omega_i \phi_i^{2}}{2}\Bigr] \exp\Bigl[-\frac{(\vec{\lambda} \oslash \sqrt{\vec{\tau}})^\top (\vec{\lambda} \oslash \sqrt{\vec{\tau}})}{2}\Bigr],
\label{fMax:eqn}
\end{equation}
}where $\kappa_{i} = y_i - \frac{1}{2}$ and $y_i \in \{0, 1\}$. And the operator $\oslash$ represents element-wise division between the two vectors $\vec{\lambda}$ \& $\vec{\tau}$.

The first exponential term of the right hand side (RHS) of (\ref{fMax:eqn}) is the joint conditional likelihood of the binary outcome $y_i \in \{0, 1\}$, denoted as $P(y_i|\vec{\lambda}, A_i)$ and the conditional likelihood of the P\'olya-Gamma distributed variable (from data augmentation) denoted as $P(\omega_i|\vec{\lambda}, A_i)$. The second exponential term of the RHS of (\ref{fMax:eqn}) is the {\it Gaussian} prior on $\vec{\lambda}$ with variance $\vec{\tau}$.
The stochastic natural gradient ascent-style updates for $\vec{\lambda}$ at step $t$ is given as:
\begin{equation} 
\vec{\lambda}_{(t)} = \vec{\lambda}_{(t-1)} + \gamma_t.\mathcal{I}(\vec{\lambda}_{(t-1)})^{-1}\mathop{\mathbb{E}}_{y_i, \omega_i: i \in I_t} \Bigl[\nabla_{\vec{\lambda}} \log[g(\vec{\lambda})]\Bigr], \label{sgd2:eqn}
\end{equation}
where $\gamma_t$ in (\ref{sgd2:eqn}) is given by the following equation:
\begin{equation*}
%\gamma_t = \max\Bigl[\frac{1}{(\tau_{p} + t)^{\theta}}, \frac{Q}{\eta_{0} + Wt}\Bigr].
\gamma_t = \frac{1}{(\tau_{p} + t)^{\theta}}.
\end{equation*}
$\mathcal{I}(\vec{\lambda}_{(t-1)})^{-1}$ in (\ref{sgd2:eqn}) denotes the inverse of the partial Fisher information matrix whose size is $R \times R$ since the second order derivatives are computed only w.r.t. $\vec{\lambda}$.

Note that the joint likelihood term in (\ref{fMax:eqn}) is un-normalized. The partial Fisher information matrix is computed only w.r.t. the data i.e., $y_i$. To do this, we first compute the expectation w.r.t. the data augmented variable $\omega_i$. The resulting equation is a Logistic function from the following identity:
\begin{equation}
\frac{\exp[\phi_i]^{y_i}}{1 + \exp[\phi_i]} = \frac{\exp[\kappa_i\phi_i]}{2}\int_0^{\infty} \exp[-\frac{\omega_i\phi_i^2}{2}]p(\omega_i)\mathrm{d}\omega_i.
\label{cosh:eqn}
\end{equation}
There is a closed-form solution for the integral in (\ref{cosh:eqn}), hence we obtain:
\begin{equation}
\frac{\exp[\phi_i]^{y_i}}{1 + \exp[\phi_i]} = \frac{\exp[\kappa_i\phi_i]}{2}\frac{1}{\mathrm{cosh}[\frac{\phi_i}{2}]}.
\label{cosh_2:eqn}
\end{equation}
Equation (\ref{cosh_2:eqn}) is a normalized likelihood for each $i \in I_t$ and let it be denoted by $\mathcal{L}_i$. Using the definition of hyperbolic cosine, (\ref{cosh_2:eqn}) becomes:
\begin{equation*}
\mathcal{L}_i = \frac{\exp \Bigl[\kappa_i\phi_i\Bigr]}{\exp \Bigl[-\frac{\phi_i}{2}\Bigr] + \exp \Bigl[\frac{\phi_i}{2}\Bigr]}.
\end{equation*}
To this end, the partial Fisher Information with respect to $\vec{\lambda}$ for $\mathcal{L}_i$ and $i \in I_t$ is given by:
\begin{equation*}
I_{\mathcal{L}}(\vec{\lambda}) = -\mathop{\mathbb{E}}_{y_i: i \in I_t} \Bigl[\sum_{i \in I_t} \frac{\partial^{2} log[\mathcal L_i]}{\partial \vec{\lambda}^{2}}\Bigr] = [\boldsymbol{A}_{I_t}^\top  \boldsymbol{N}_{I_t} \boldsymbol{A}_{I_t}],
\end{equation*}
where $\boldsymbol{A}_{I_t}$ denotes the matrix whose rows are $A_i$ for $i \in I_t$ and $\boldsymbol{N}_{I_t}$ denotes the diagonal matrix whose diagonal elements are $N_{ii}$; where:
\begin{equation*}
N_{ii} = \frac{1}{\Bigl[\exp[-\frac{\phi_i}{2}] + \exp [\frac{\phi_i}{2}]\Bigr]^2}.
\end{equation*}
The prior term is accounted by considering its variance as a conditioner, hence the conditioned partial Fisher Information matrix for the parameter $\vec{\lambda}$ at step $t$ is given by:
\begin{equation}\label{eq:fisher_information_lambda}
\mathcal{I}(\vec{\lambda}_{(t-1)}) = [\boldsymbol{A}_{I_t}^\top  \boldsymbol{N}_{I_t} \boldsymbol{A}_{I_t}] + \text{diag}[\vec{\tau}_{(t-1)}]^{-1},
\end{equation}
where diag$[\vec{\tau}_{(t-1)}]^{-1}$ denotes inverse of a diagonal matrix whose diagonal is $\vec{\tau}_{(t-1)}$.

\subsection*{Partial Fisher information matrix w.r.t. $\vec{\beta}^{(k)}$ for mode $k$}

The function to be maximized w.r.t. $\vec{\beta}^{(k)}$ is:
\begin{equation}
\begin{split}
F(\vec{\beta}^{(k)})=\Bigl[
\sum_{m = 1}^{M} [z_{m}^{(k)}\frac{\psi_{m}^{(k)}}{2} - \frac{\nu_{m}^{(k)} \psi_{m}{^{(k)}}^{2}}{2}]
- \frac{(\vec{\beta}^{(k)} \oslash \vec{\rho}^{(k)})^\top (\vec{\beta}^{(k)} \oslash \vec{\rho}^{(k)})}{2}]
\Bigr],
\end{split}
\label{fBeta:eqn}
\end{equation}
where the operator $\oslash$ represents element-wise division between the two vectors $\vec{\beta}^{(k)}$ \& $\vec{\rho}^{(k)}$.

Let $\tilde{\boldsymbol{U}}_M^{(k)}$ be the matrix whose rows are $\tilde{\vec{u}}_{i_k=m}^{(k)}$ for $m \in [1,M]$ for mode $k$ with target information.
We compute the partial Fisher information matrix $I(\vec{\beta}^{(k)}_{(t-1)})$ similar to $\vec{\lambda}$ as:
\begin{equation*}
\mathcal{I}(\vec{\beta}^{(k)}) = [\tilde{\boldsymbol{U}}_{M}^{(k)}]^\top \boldsymbol{N}_M \tilde{\boldsymbol{U}}_{M}^{(k)} + \text{diag}[\vec{\rho}{^{(k)}}^{2}]^{-1},
\end{equation*}
where the $\boldsymbol{N}_M$ is the diagonal matrix whose diagonal elements are $N_{mm}$ where:
\begin{equation*}
N_{m,m} = \frac{1}{\Bigl[\exp[-\frac{\psi_{m}^{(k)}}{2}] + \exp [\frac{\psi_{m}^{(k)}}{2}]\Bigr]^2}.
\end{equation*}

\subsection*{Partial Fisher information matrix w.r.t. $\vec{u}_{i_k=n}^{(k)}$ for mode $k$: Without Target Information}

The function to be maximized w.r.t. factor $\vec{u}_{i_k=n}^{(k)}$ is:
{\small
\begin{equation}
\begin{array}{l}
F(\vec{u}_{i_k=n}^{(k)})=\Bigl[ 
\sum_{i \in I_t: i_k = n} [\frac{\phi_i}{2} - \frac{\omega_i \phi_i^{2}}{2}]
- \frac{(\vec{u}_{i_k=n}^{(k)} \oslash \vec{\mu}_{i_k=n})^\top (\vec{u}_{i_k=n}^{(k)} \oslash \vec{\mu}_{i_k = n})}{2} + (\vec{\zeta}_{i_k=n})^{\top}\vec{u}_{i_k=n}^{(k)}
\Bigr],
\end{array}
\label{fUeq:eqn}
\end{equation}
}where $\vec{\zeta}_{i_k=n}$ is the vector of Lagrange multipliers for the non-negativity constraint on the $\vec{u}_{i_k=n}^{(k)}$. And the operator $\oslash$ represents element-wise division between the two vectors $\vec{u}_{i_k=n}^{(k)}$ \& $\vec{\mu}_{i_k = n}$.

Similarly we can compute the partial Fisher information for $\vec{u}_{i_k=n}^{(k)}$ corresponding to element $n$ in mode $k$ as:
\begin{equation*}
\mathcal{I}(\vec{u}_{i_k=n}^{(k)}{_{(t-1)}}) = [\boldsymbol{C}_{n}^{(k)}]^\top \boldsymbol{N}_{n} \boldsymbol{C}_{n}^{(k)} + \text{diag}[\vec{\mu}_{i_k=n}^{2}]^{-1},
\end{equation*}
where $\boldsymbol{C}_n^{(k)}$ is a matrix whose rows are $\vec{C}_{i_k = n}$ for $i \in I_t$ and $\boldsymbol{N}_{n}$ is the diagonal matrix whose diagonal elements $N_{i:i_k = n}$ is given by:
\begin{equation*}
N_{i:i_k=n} = \frac{1}{\Bigl[\exp[-\frac{\phi_{i:i_k=n}}{2}] + \exp[\frac{\phi_{i:i_k=n}}{2}]\Bigr]^2}.
\end{equation*}

\subsection*{Partial Fisher information matrix w.r.t. $\vec{u}_{i_k=n}^{(k)}$ for mode $k$: With Binary Target Information}

The function to be maximized w.r.t. factor $\vec{u}_{i_k=n}^{(k)}$ is:
\begin{equation}
\begin{array}{lll}
F(\vec{u}_{i_k=n}^{(k)}) & = & \Bigl[
\sum_{i \in I_t: i_k = n} [\frac{\phi_i}{2} - \frac{\omega_i \phi_i^{2}}{2}] 
 - \frac{(\vec{u}_{i_k=n}^{(k)} \oslash \vec{\mu}_{i_k=n})^\top (\vec{u}_{i_k=n}^{(k)} \oslash \vec{\mu}_{i_k = n})}{2}\\
&  & \quad + [z_{n}^{(k)}.\frac{(\vec{u}_{i_k=n}^{(k)})^{\top}\vec{\hat{\beta_l}}^{(k)}}{2} - \frac{\nu_{n}^{(k)}[\beta_{0}^{(k)} + (\vec{u}_{i_k=n}^{(k)})^{\top}\vec{\hat{\beta}}^{(k)}]^{2}}{2}]
\Bigr],
\end{array}
\label{fUeqSI:eqn}
\end{equation}
where the operator $\oslash$ represents element-wise division between the two vectors $\vec{u}_{i_k=n}^{(k)}$ \& $\vec{\mu}_{i_k = n}$. Note that (\ref{fUeqSI:eqn}) does not have the non-negativity constraint since we are training a Logistic model using the binary target information to detect abusive entities.

Similarly we can compute the partial Fisher information for $\vec{u}_{i_k=n}^{(k)}$ corresponding to element $n$ in mode $k$ as:
{\small
\begin{equation*}
\begin{array}{l}
\mathcal{I}(\vec{u}_{i_k=n}^{(k)}{_{(t-1)}}) = [\boldsymbol{C}_{n}^{(k)}]^\top \boldsymbol{N}_{n} \boldsymbol{C}_{n}^{(k)} + [\vec{\beta}^{(k)}]^\top N_{m=n} \vec{\beta}^{(k)}  +  \text{diag}[\vec{\mu}_{i_k=n}^{2}]^{-1},
\end{array}
\end{equation*}}
where $N_{m=n}$ is:
\begin{equation*}
N_{m=n} = \frac{1}{\Bigl[\exp[-\frac{\psi_{m=n}^{(k)}}{2}] + \exp[\frac{\psi_{m=n}^{(k)}}{2}]\Bigr]^2}.
\end{equation*}

\section{Computation of the gradient}\label{supp_sec:gradient}

\subsection*{Gradient w.r.t. $\vec{\lambda}$}
To update $\vec{\lambda}$, we compute the gradient of the natural logarithm of (\ref{fMax:eqn}) as:
{\small
\begin{equation}\label{grad_lambda:eqn}
\begin{array}{l}
\mathop{\mathbb{E}}_{y_i, \omega_i: i \in I_t} \Bigl[\nabla_{\vec{\lambda}} \log[g(\vec{\lambda})]\Bigr] = \mathop{\mathbb{E}}_{y_i, \omega_i: i \in I_t} \Bigl[\Bigl[\sum_{i \in I_t} \kappa_iA_i - A^\top_i \omega_i (A_i \vec{\lambda}) \Bigr] - \text{diag}[\vec{\lambda} \oslash \vec{\tau}] \Bigr].
\end{array}
\end{equation}
}

Separating out terms independent of $y_i$ \& $\omega_i$ and replacing with matrix operations where applicable; the RHS of (\ref{grad_lambda:eqn}) becomes:
\begin{equation*}
\sum_{i \in I_t} \mathop{\mathbb{E}}_{y_i} \Bigl[\kappa_i \Bigr]A_i - \Bigl[\boldsymbol{A}_{I_t}^\top \hat{\boldsymbol{\omega}}_{I_t} \boldsymbol{A}_{I_t}  + \text{diag}[\vec{\tau}]^{-1} \Bigr]\vec{\lambda},
\end{equation*}
where $\hat{\boldsymbol{\omega}}_{I_t}$ is a diagonal matrix whose diagonal elements are $\hat{\omega}_i$, where:
\begin{equation*}
\hat{\omega}_i = \mathop{\mathbb{E}}[\omega_i] = \frac{\text{tanh}(\frac{\phi_i}{2})}{2\phi_i}.
\end{equation*}

Now $\kappa_i = -0.5$ if  $y_i = 0$ and $\kappa_i = 0.5$ if $y_i = 1$. Hence:
\begin{equation*}
\begin{array}{l}
\mathop{\mathbb{E}}_{y_i} \Bigl[\kappa_i | y_i = 0  \Bigr] = -0.5\frac{\exp(-\frac{\phi_i}{2})}{\exp(-\frac{\phi_i}{2}) + \exp(\frac{\phi_i}{2})} \quad \text{and}\\
\mathop{\mathbb{E}}_{y_i} \Bigl[\kappa_i | y_i = 1  \Bigr] = 0.5\frac{\exp(\frac{\phi_i}{2})}{\exp(-\frac{\phi_i}{2}) + \exp(\frac{\phi_i}{2})}.
\end{array}
\end{equation*}

\subsection*{Gradient w.r.t. $\vec{\beta}^{(k)}$}
To update $\vec{\beta}^{(k)}$, we compute the gradient of (\ref{fBeta:eqn}) as:
\small{
\begin{equation}\label{grad_beta:eqn}
\begin{array}{lll}
\mathop{\mathbb{E}}_{z_{m}^{(k)}, \nu_{m}^{(k)}: m \in [1, M]} \Bigl[\nabla_{\vec{\beta}^{(k)}} F(\vec{\beta}^{(k)})\Bigr] & = & \mathop{\mathbb{E}}_{z_{m}^{(k)}, \nu_{m}^{(k)}: m \in [1,M]} \Bigl[ 
\sum_{m = 1}^{M} [z_{m}^{(k)} \frac{\tilde{\vec{u}}^{(k)}_{i_k = m}}{2}  \\
 & & - ({\tilde{\vec{u}}^{(k)}_{i_k = m}}^\top \nu_{m}^{(k)} \tilde{\vec{u}}^{(k)}_{i_k = m})\vec{\beta}^{(k)}]
- \text{diag}[\vec{\beta}_l \oslash \vec{\rho}_l]
\Bigr].
\end{array}
\end{equation}}Separating out terms independent of $z_{m}^{(k)}$ \& $\nu_{m}^{(k)}$ and replacing with matrix operations where applicable; the RHS of (\ref{grad_beta:eqn}) becomes:
\begin{equation*}
\sum_{m = 1}^{M} \mathop{\mathbb{E}}_{z_{m}^{(k)}} \Bigl[z_{m}^{(k)} \Bigr]\frac{\tilde{\vec{u}}^{(k)}_{i_k = m}}{2} - \Bigl[{\tilde{\boldsymbol{U}}_{M}^{(k)\top}}  \hat{\boldsymbol{\nu}}_M^{(k)} \tilde{\boldsymbol{U}}_{M}^{(k)} + \text{diag}[\vec{\rho}{^{(k)}}^2]^{-1} \Bigr]\vec{\beta}_l,
\end{equation*}
where $\hat{\boldsymbol{\nu}}_M^{(k)}$ is a diagonal matrix whose diagonal elements are $\hat{\nu}_{m}^{(k)}$, where:
\begin{equation*}
\hat{\nu}_{m}^{(k)} = \mathop{\mathbb{E}}[\nu_{m}^{(k)}] = \frac{\text{tanh}(\frac{\psi_m^{(k)}}{2})}{2\psi_m^{(k)}}.
\end{equation*}

Subsequently:
\begin{equation*}
\begin{array}{l}
\mathop{\mathbb{E}}_{z_{m}^{(k)}} \Bigl[z_{m}^{(k)} | label_{m}^{(k)} = 0  \Bigr] = -\frac{\exp(-\frac{\psi_{m}^{(k)}}{2})}{\exp(-\frac{\psi_{m}^{(k)}}{2}) + \exp(\frac{\psi_{m}^{(k)}}{2})} \quad \text{and}\\
\mathop{\mathbb{E}}_{z_{m}^{(k)}} \Bigl[z_{m}^{(k)} | label_{m}^{(k)} = 1  \Bigr] = \frac{\exp(\frac{\psi_{m}^{(k)}}{2})}{\exp(-\frac{\psi_{m}^{(k)}}{2}) + \exp(\frac{\psi_{m}^{(k)}}{2})}.
\end{array}
\end{equation*}

\subsection*{Gradient w.r.t. $\vec{u}_{i_k=n}^{(k)}$: Without Target Information}
To update $\vec{u}_{i_k=n}^{(k)}$, we compute the gradient of (\ref{fUeq:eqn}) as:
\begin{equation*}
\sum_{i \in I_t:i_k=n} \mathop{\mathbb{E}}_{y_i} \Bigl[\kappa_i \Bigr]C_{i_k=n}^{(k)} - \Bigl[ {\boldsymbol{C}_{n}^{(k)}}^\top \hat{\boldsymbol{\omega}}_n \boldsymbol{C}_{n}^{(k)}  + \text{diag}[\vec{\mu}_{i_k=n}^{2}]^{-1} \Bigr]\vec{u}_{i_k=n}^{(k)},
\end{equation*}
where $\hat{\boldsymbol{\omega}}_{n}$ is a diagonal matrix whose diagonal elements are $\hat{\omega}_{i \in I_t:i_k=n}$.

\subsection*{Gradient w.r.t. $\vec{u}_{i_k=n}^{(k)}$: With Binary Target Information}

To update $\vec{u}_{i_k=n}^{(k)}$, we compute the gradient of (\ref{fUeqSI:eqn}) as:
\begin{equation*}
\begin{array}{lll}
\sum_{i \in I_t:i_k=n} \mathop{\mathbb{E}}_{y_i} \Bigl[\kappa_i \Bigr]C_{i_k=n}^{(k)} + \mathop{\mathbb{E}}_{z_{m=n}^{(k)}} \Bigl[z_{m=n}^{(k)} \Bigr]\frac{\hat{\vec{\beta}}^{(k)}}{2} 
- \Bigl[{\boldsymbol{C}_{n}^{(k)}}^\top \hat{\boldsymbol{\omega}}_n \boldsymbol{C}_{n}^{(k)}  + \text{diag}[\vec{\mu}_{i_k=n}^{2}]^{-1} - ({\hat{\vec{\beta}}^{(k)}}^\top\hat{\nu}_{m=n}^{(k)}\hat{\vec{\beta}}^{(k)}) \Bigr]\vec{u}_{i_k=n}^{(k)}.
\end{array}
\end{equation*}